\pdfoutput=1

\documentclass[11pt]{article}

\usepackage[preprint]{acl}

\usepackage{times}
\usepackage{latexsym}

\usepackage[T1]{fontenc}

\usepackage[utf8]{inputenc}

\usepackage{microtype}

\usepackage{inconsolata}

\usepackage{graphicx}

\usepackage{booktabs}
\usepackage{amsmath}
\usepackage{booktabs}
\usepackage[table]{xcolor}  
\usepackage[np]{numprint}
\usepackage{colortbl}
\usepackage{multirow}
\usepackage[font=small, labelfont=bf]{caption}
\usepackage{url}
\usepackage[normalem]{ulem}
\usepackage{xcolor}
\usepackage{orcidlink}

\newcommand{\OMorcid}{\orcidlink{0000-0001-8230-1453}}
\newcommand{\SZorcid}{\orcidlink{0000-0002-1384-1218}}

\newcommand{\posneg}[1]{%
  \begingroup
  \ifdim #1pt<0pt
    \color{red}%
  \else
    \color{green!50!black}%
  \fi
  \ensuremath{#1}%
  \endgroup
}

\definecolor{lightgray}{gray}{0.9}

\title{The Frequency Confound in Language-Model Surprisal and Metaphor Novelty}

\author{
    Omar Momen\OMorcid \and
    Sina Zarrieß\SZorcid \\
    CRC 1646 – Linguistic Creativity in Communication \\ 
    Bielefeld University, Germany \\
    \texttt{\{omar.hassan, sina.zarriess\}@uni-bielefeld.de}
}

\begin{document}
\maketitle
\begin{abstract}
Language-model (LM) surprisal is widely used as a proxy for contextual predictability and has been reported to correlate with metaphor novelty judgments. However, surprisal is tightly intertwined with lexical frequency. We explore this interaction on metaphor novelty ratings using two different word frequency measures. We analyse surprisal estimates from eight Pythia model sizes and 154 training checkpoints. Across settings, word frequency is a stronger predictor of metaphor novelty than surprisal. Across training stages, the surprisal--novelty association peaks at an early stage and then falls again, mirroring a similarly timed increase in the surprisal--frequency association. These results suggest that the often-reported optimal LM surprisal settings may incorrectly associate contextual predictability with metaphor novelty and processing difficulty, whereas lexical frequency may be the major underlying factor. The experiment resources are publicly available.\footnote{Data and code: \url{https://github.com/OmarMomen14/surprisal_frequency_novelty}}
\end{abstract}

\section{Introduction}
Conceptual Metaphor Theory~\cite{cmt1980} argues that a metaphor is not only a linguistic phenomenon but also a cognitive mechanism through which abstract concepts are mapped into more concrete domains. Many of such metaphorical mappings are highly conventionalised, e.g.,  ARGUMENT $\rightarrow$ WAR, TIME $\rightarrow$ MONEY, or LIFE $\rightarrow$ JOURNEY.

But metaphorical mappings can vary in their degree of novelty: whereas we ``live by'' many conventionalised metaphors in everyday language,
novel metaphors can stand out as instances of linguistic creativity. For example, the metaphorical mappings of WATER in the sentence: ``The arrested water shone and danced.''\footnote{Example from the VU Amsterdam Metaphor
Corpus (VUAMC)~\cite{vuamc}, and the sentence is traced back to the novel ``Still Life'' by A. S. Byatt, 1985.} are considered relatively novel. Novel (creative) metaphors introduce less familiar mappings and require greater interpretive effort to understand~\cite{lai2009,cardillo_etal_2012_novel_to_familiar,philip_2016_conventional_novel}. Commonly, metaphor novelty is measured by human ratings~\cite{do-dinh-etal-2018-weeding}.

Recent studies have shown that metaphor novelty ratings by humans correlate significantly with both lexical frequency~\cite{do-dinh-etal-2018-weeding,reimann-scheffler-2024-metaphor} and LM surprisal~\cite{momen-etal-2026-surprisal}. Interestingly, \citet{momen-etal-2026-surprisal} report that, within the same model family, surprisal estimates from smaller LMs consistently correlate more strongly with metaphor novelty ratings than those from larger variants. This mirrors the inverse scaling effect observed in studies of reading times, where smaller models often provide better fits to human reading behaviour~\citep{oh-schuler-2023-transformer,oh-schuler-2023-surprisal,de-varda-marelli-2023-scaling}.

Lexical frequency has been proposed as a potential explanation for the inverse scaling effect in reading time studies. \citet{oh-etal-2024-frequency} argue that larger LMs are better than humans at predicting low-frequency words, which weakens the correlation between their surprisal estimates and human reading-time data. Also, \citet{opedal-etal-2024-role} report that non-contextual frequency has a stronger effect on predicting reading time than surprisal.

Surprisal Theory \citep{hale_2001,levy2008expectation} is a common approach to studying processing difficulties in humans and LMs. The theory predicts that processing difficulty increases as a word becomes less predictable in context. Recently, the consistent inverse scaling pattern reported in studies of different phenomena such as reading time or metaphor novelty~\citep{oh-schuler-2023-transformer,de-varda-marelli-2023-scaling,momen-etal-2026-surprisal} has motivated an essential assumption that smaller LMs or shorter pretraining provide more human-aligned surprisal estimates. However, whether this pattern reflects genuinely more human-like contextual prediction or confounding lexical properties remains unclear.

In this paper, we analyse the associations between the three quantities: ``Metaphor Novelty Ratings'', ``Lexical Frequency'' and ``LM Surprisal''. We propose two different methods for computing lexical frequency. Then, we compute LM surprisal estimates across (i) eight Pythia model sizes and (ii) 154 pretraining checkpoints. Our results show that word frequency is a substantially stronger predictor of novelty than surprisal across settings, and that the configurations where surprisal performs ``best'' are also those where it aligns most closely with frequency. We therefore caution against interpreting strong early-training or small-model surprisal--novelty associations as straightforward support for surprisal-based accounts of metaphor novelty, and we argue that progress requires clearer theoretical and methodological separation between contextual predictability and lexical-frequency effects.

\section{Data \& Methods}

This Section describes our experimental set-up to measure the association between word frequency and metaphor novelty on the one hand and LM surprisal on the other hand.

\subsection{Dataset}
We base our study on the VU Amsterdam Metaphor Corpus (VUAMC)~\cite{vuamc}. Every word in VUAMC is annotated as either a metaphoric word or not. Building on VUAMC, \citet{do-dinh-etal-2018-weeding} collected crowd-sourced metaphor novelty ratings for the 15,155 metaphoric content words in VUAMC and converted them into continuous scores in the range (-1, +1), where -1 denotes the most conventional and +1 the most novel. They additionally binarised these scores using a 0.5 threshold, resulting in 353 metaphors labelled as novel out of the 15,155 content metaphoric words (see Appendix~\ref{sec: app_data}). In our experiment, we use these 15,155 instances, each consisting of a single sentence, a target word within the sentence (a content metaphoric word), and an associated metaphor novelty score.

\subsection{Model Suite}
To examine the effects of model scale and pretraining progress (data/steps), we use the Pythia model suite~\cite{biderman2023pythiasuiteanalyzinglarge}. Pythia consists of decoder-only causal LMs at 8 sizes (70M--12B parameters), all trained on the same 300B-token pretraining corpus \footnote{The Pile~\cite{gao2020pile800gbdatasetdiverse}} in the same order. For each model, Pythia provides 154 intermediate checkpoints saved every 1{,}000 training steps (corresponds to additional $\approx$2M tokens seen during these steps), and denser early checkpoints at steps $\{1, 2, 4, 8, 16, 32, 64, 128, 256, 512\}$. The exact pretraining sequences seen at each checkpoint can be reconstructed using available scripts.

\subsection{Surprisal}

For causal LMs, surprisal, computed for a word \(w_i\) within a sequence \(W_{0:n}\) is \(\mathrm{Surprisal}(w_i)=-\log p(w_i \mid w_{<i})\)\footnote{We use \(\log\) of base \(e\) for all \textit{log} computations in our study.}. In our experiment, we measure word-level surprisal for metaphoric target word(s) in their sentence-level context by running an independent, teacher-forced forward pass per sentence and recording the target word surprisal. To map token probabilities to a target word, we locate the word’s character offsets in the sentence and sum token-level surprisals over the minimal token span in the sequence's tokenisation that covers these offsets. We additionally apply word-probability corrections for leading-whitespace tokenisation confounds~\cite{pimentel-meister-2024-compute}, and we prepend a BOS token so surprisal is defined even when the target is the first word in a sentence.

\subsection{Word Frequency}
Estimating word frequency is non-trivial because frequency is not directly observed as a property of a word, but estimated from a reference corpus. Different corpora approximate different kinds of language exposure, and there is no single corpus that can be assumed to represent the linguistic experience of all speakers. To tackle this problem, we utilise two different estimates of word frequency reflecting both human and LM lexical memory. In both estimates, we compute the \emph{negative log frequency} of each target metaphoric word for easier numerical comparability with surprisal.\\

\begin{enumerate}
    \item \textbf{Negative Log Frequency in General Language Use:} We compute the negative log frequency of each target metaphoric word using the Python library \textit{wordfreq}\footnote{\url{https://pypi.org/project/wordfreq/}}, which provides corpus-independent frequency estimates aggregated from multiple large-scale sources, rather than deriving counts from a single corpus. We treat this as an estimate of word frequency for an ``average English speaker'', and hereafter denote it as \textbf{NLF-Human}.

    \item \textbf{Negative Log Frequency in Pythia's Pretraining Data:}
    For each target metaphoric word, we tokenise its sentence using Pythia's tokeniser and, at each checkpoint, we count occurrences of the target word's subtoken sequence in the pretraining tokens seen up to that checkpoint. We treat the negative log of these frequencies as an LM checkpoint-specific estimate of word frequency. Hereafter, we denote this estimate as \textbf{NLF-LM}.
\end{enumerate}

\subsection{Experiment}
We compute surprisal using all 8 Pythia model sizes and at each of the 154 checkpoints of Pythia-70M. We likewise compute NLF-LM at each of these checkpoints, and NLF-Human (once) per metaphoric word.
We quantify surprisal--novelty and frequency--novelty associations using Spearman’s ($\rho$) and Pearson’s ($r$)  correlation coefficients, and we report the Area Under the ROC Curve (AUC) as an estimate of discriminating binarised novel metaphors using surprisal or frequency as a predictor. Additionally, Surprisal--frequency associations are estimated using Spearman’s ($\rho$) correlation.

\section{Results}
\begin{figure}[!ht]
  \centering
  \includegraphics[width=\linewidth]{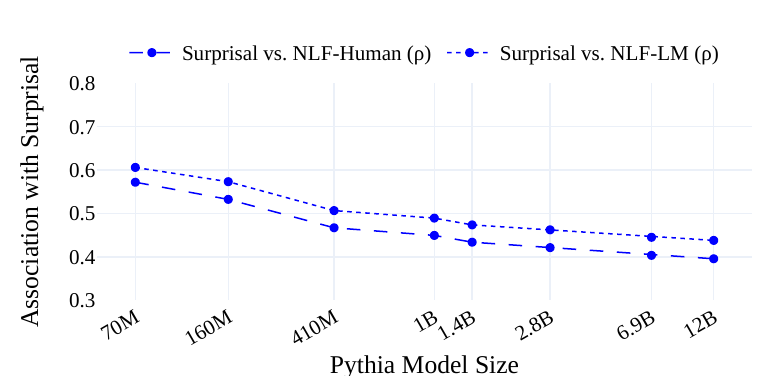}
  \caption{Effect of model scale on correlation between Surprisal and Frequency. NLF-Human (\textbf{dash}); and NLF-LM (\textbf{dots}).}
  \label{fig: model_size surp effect}
\end{figure}

\begin{figure}[!ht]
  \centering
  \includegraphics[width=\linewidth]{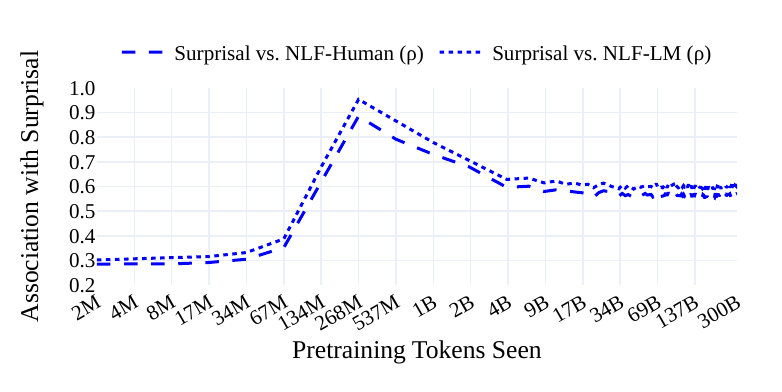}
  \caption{Effect of pretraining data/steps of Pythia-70M on correlation between Surprisal and Frequency. NLF-Human (\textbf{dash}); and NLF-LM (\textbf{dots}).}
  \label{fig: model_steps surp effect}
\end{figure}

Figures~\ref{fig: model_size effect}--\ref{fig: model_steps surp effect} visually illustrate the results across model sizes and checkpoints.
All detailed numerical results are provided in Appendix~\ref{sec:appendix_res}. 

\subsection{Associations with Metaphor Novelty}
\begin{figure*}[!ht]
  \centering
  \includegraphics[width=\linewidth]{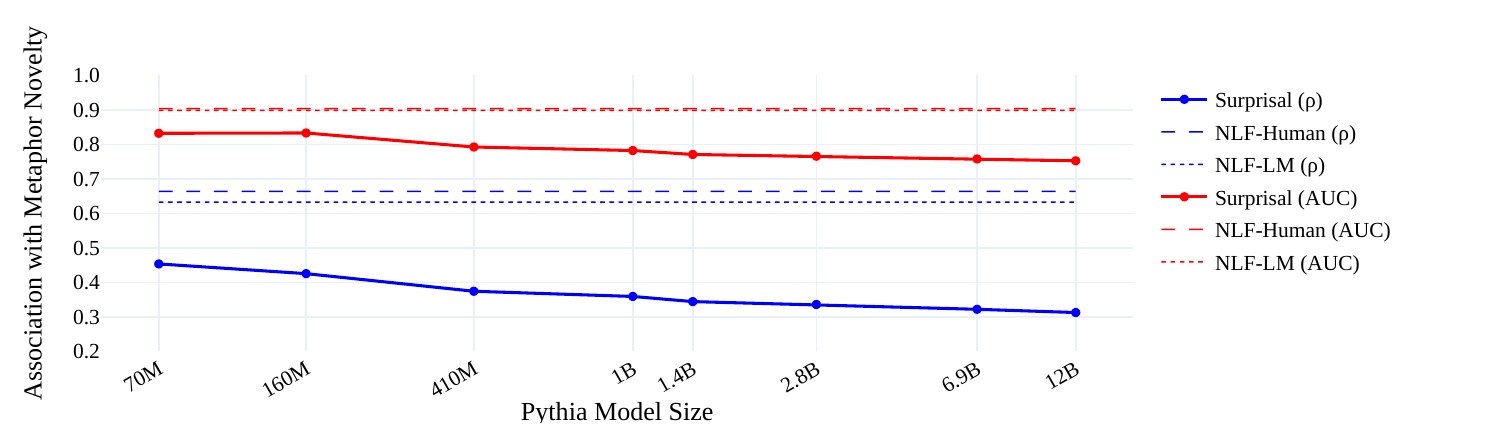}
  \caption{Effect of model size on associations between Metaphor Novelty Scores and Surprisal (\textbf{solid}); Negative Log Word Frequency in general language use (NLF-Human) (\textbf{dash}); and Negative Log Word Frequency in Pythia's pretraining data (NLF-LM) (\textbf{dots}). Blue lines track Spearman correlation, and red lines track AUC to detect novel metaphors ($score\geq0.5$).}
  \label{fig: model_size effect}
\end{figure*}

\paragraph{Model Scale:}
Figure~\ref{fig: model_size effect} shows surprisal--novelty and frequency--novelty association across Pythia model sizes. Here, NLF-LM is computed at the final checkpoint (300B tokens), and is therefore identical across sizes. Overall, frequency has a clearly stronger association with novelty than surprisal, with NLF-Human yielding slightly higher estimates ($\rho=.66, r=.66, AUC=.90$) than NLF-LM ($\rho=.63, r=.60, AUC=.90$). We also observe a consistent negative effect of model scale on the surprisal--novelty association.

\begin{figure*}[!ht]
  \centering
  \includegraphics[width=\linewidth]{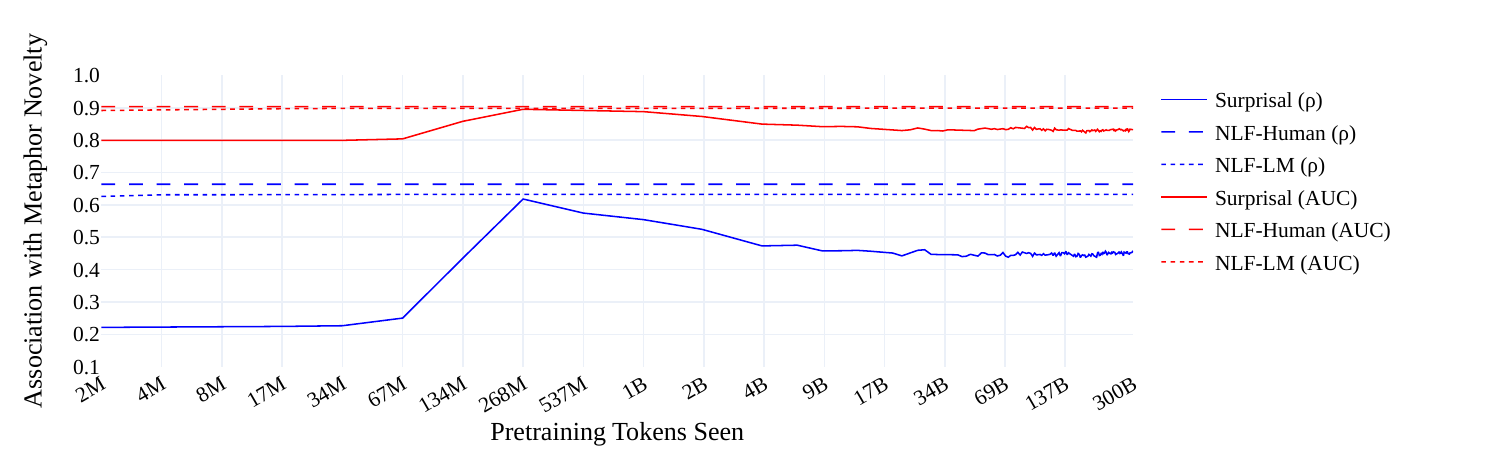}
  \caption{Effect of pretraining data/steps for Pythia-70M on associations between Metaphor Novelty Scores and Surprisal (\textbf{solid}); Negative Log Word Frequency in general language use (NLF-Human) (\textbf{dash}); and Negative Log Word Frequency in Pythia's pretraining data (NLF-LM) (\textbf{dots}). Blue lines track Spearman correlation, and red lines track AUC.}
  \label{fig: model_steps effect}
\end{figure*}

\paragraph{Pretraining Progress:}
Figure~\ref{fig: model_steps effect} reports associations across the 154 checkpoints of Pythia-70M. Despite the change of NLF-LM across checkpoints, NLF-LM--novelty association remains almost the same across checkpoints, except for a small deviation at the earliest steps (1--4). In contrast, the surprisal--novelty association is very weak in the first checkpoints, then rises sharply after 64 training steps (134M tokens), and peaks after 128 steps (268M tokens), where it approaches the frequency estimates ($\rho=.62, AUC=.90$). After this peak, the surprisal--novelty association converges to ($\rho=.45, AUC=.83$). Yet, surprisal never reaches the same strength of association with novelty as frequency does.

\subsection{Correlations between Surprisal and Frequency}

\paragraph{Model Scale:}
Figure~\ref{fig: model_size surp effect} reports frequency--surprisal Spearman correlations across Pythia model sizes. Here, both NLF-Human and NLF-LM are fixed across sizes, while surprisal changes. Across sizes, surprisal shows moderate correlations with both NLF estimates ($\rho \in [.40, .61]$). Correlations decrease with model size, mirroring the negative scale effect observed for associations with novelty (Figure~\ref{fig: model_size effect}), suggesting that larger models’ surprisal diverges more from frequency estimates. Overall, surprisal here correlates slightly more with NLF-LM (max $\rho=.61$) than with NLF-Human (max $\rho=.57$).

\paragraph{Pretraining Progress:}
Figure~\ref{fig: model_steps surp effect} reports frequency--surprisal correlations across the 154 checkpoints of Pythia-70M. Here, NLF-Human is fixed, while NLF-LM and surprisal vary with checkpoint. The correlation pattern of surprisal to frequency closely matches the trends of surprisal's association to novelty in Figure~\ref{fig: model_steps effect}: here surprisal has a weak correlation to frequency at the earliest checkpoints, then the correlation rise sharply after 64 training steps (134M tokens), and peak after 128 steps (268M tokens), reaching $\rho=.95$ with NLF-LM and $\rho=.89$ with NLF-Human. Correlations then gradually converge across subsequent checkpoints, with moderate strength $\rho\approx.60$.

 \section{Discussion}

\textbf{Word Frequency:}
Our results agree with previous work~\cite{do-dinh-etal-2018-weeding,reimann-scheffler-2024-metaphor} showing that lexical frequency is strongly associated with metaphor novelty scores. Additionally, we show that frequency--novelty association is substantially stronger than surprisal--novelty associations across all LM settings in our experiment. Notably, frequency correlates more strongly with novelty than it does with surprisal itself. We further observe that NLF-Human aligns slightly more with novelty (human-based) than NLF-LM, whereas NLF-LM aligns slightly more with surprisal (LM-based) than NLF-Human. Overall, however, differences between the two frequency estimates are small and do not alter overall trends, suggesting that estimating frequencies from relatively small amounts of data is sufficient (at least for our task) when large-scale estimation is expensive.

\paragraph{Inverse Scale Effect:}
The surprisal--novelty association decreases with model size, replicating prior results on the same dataset across other model families 
~\cite{momen-etal-2026-surprisal}, and on datasets of reading time~\cite{oh-schuler-2023-surprisal}. We additionally show that the same negative scaling trend holds for correlations between surprisal and frequency: as model size increases, surprisal becomes less aligned with frequency.

\paragraph{Pretraining Amount:}
The strongest association between surprisal and novelty, and between surprisal and frequency, occurs early---after 128 pretraining updates for Pythia-70M ($\approx$268M tokens seen)---and additional training weakens this association. A qualitatively similar non-monotonic effect has been reported for reading times: surprisal predicts reading time best at an intermediate pretraining amount (about 2B tokens), after which further pretraining reduces predictive power~\cite{oh-schuler-2023-transformer}. The close similarity between the checkpoint trends for surprisal--novelty (Figure~\ref{fig: model_steps effect}) and surprisal--frequency (Figure~\ref{fig: model_steps surp effect}) highlights the extent to which word frequency can confound surprisal-based analyses of linguistic and psycholinguistic targets.

\paragraph{Surprisal:}
Surprisal achieves its strongest association with metaphor novelty when computed from the smallest model (70M) and at a relatively early training stage (128 steps; $\approx$268M tokens). However, these are also the settings in which surprisal is most closely aligned with word frequency. We therefore caution against treating these \textit{``best''} association values as direct evidence for surprisal-as-predictability accounts of metaphor novelty (and processing difficulty in general): in these settings, surprisal may primarily reflect lexical frequency, and possibly more than contextual predictability. We think that larger models or extensive pretraining do not produce intrinsically poor estimates of predictability. And, we call for further efforts to develop more accurate human ratings that reflect clear theories of the novelty/originality dimension in creativity (and potentially processing difficulty more broadly) and to clarify how these constructs relate to surprisal-based accounts.

\section{Conclusion} 

We analysed the associations between metaphor novelty ratings, LM surprisal and lexical frequency across different model sizes and pretraining stages. The results mainly show that lexical frequency is a stronger predictor of metaphor novelty than LM surprisal. We further find that the strongest correlations between LM surprisal and metaphor novelty occur only under settings in which surprisal is also strongly correlated with lexical frequency. We conclude that relevant studies should be careful not to interpret strong surprisal--novelty associations at small or early-training LM settings as straightforward evidence for surprisal-as-predictability accounts of metaphor novelty (or processing effort).

\section*{Limitations}
Due to computational constraints, we do not compute surprisal for intermediate checkpoints of the larger Pythia models, and we restrict the checkpoint-level analysis to the smallest variant (Pythia-70M). Although this choice is consistent with our model-scale findings (smaller models yield stronger associations), evaluating intermediate checkpoints for larger models remains necessary to verify whether the observed training-dynamics trends hold across model sizes.

\section*{Acknowledgments}
This research has been funded by the Deutsche Forschungsgemeinschaft (DFG, German Research Foundation) – CRC-1646, project number 512393437, project A05.


\bibliography{custom}

\appendix

\section{Dataset Statistics}
\label{sec: app_data}

In Table~\ref{tab: data_stats}, we demonstrate the statistics of the dataset used in our experiment. In Table~\ref{tab: data_examples}, we list 16 examples from the VUAMC with their associated novelty ratings from~\citet{do-dinh-etal-2018-weeding}.
\begin{table*}[!htp]
\centering
\small
\begin{tabular}{lcc|cc}
\toprule
\textbf{Genre} & \textbf{\# Metaphors} & \textbf{L$_{sent}$}  & \textbf{Novelty Score} & \textbf{\# Novel}  \\
&  & mean | std.  & mean | std. & $>=0.5$  \\
\midrule
Fiction & 3170 & 26.0 | 16.5 &  -.005 | .271 & 94  \\
News & 4712 & 29.9 | 14.2 & .000 | .257  & 132  \\
Academic & 5499 & 34.9 | 16.0  & .003 | .239 & 102 \\ 
Conversation & 1774 & 17.5 | 15.9   & -.000 | .236 & 25  \\
\midrule
All & 15155 & 29.4 | 16.5  & .000 | .251 & 353  \\
\bottomrule
\end{tabular}
\caption{Distributions and statistics of the dataset under study. \# Metaphors is the number of metaphor words, L$_{sent}$ values are the statistics of sentences lengths in words. Novelty Score values are the normalised continuous novelty ratings, \# Novel is the number of novel metaphors (Novelty Score >= 0.5). }
\label{tab: data_stats}
\end{table*}

\begin{table*}[!htp]
\centering
\small
\begin{tabular}{c|p{9cm}|c|c}
\toprule
Genre & \multicolumn{1}{c|}{Sentence} & Novelty Score & Label \\
\midrule
\multirow{4}{*}{Fiction} &
{\fontsize{9pt}{12pt}\selectfont \textbf{1. } ‘ Tell him I am very sorry, but I must \textbf{fill} the quota. ’} &
-0.441 &
conventional
\\
\\

&{\fontsize{9pt}{12pt}\selectfont \textbf{2. } Adam might have escaped the file memories for years, \textbf{suppressed} them and jerked violently <14> by those events.} &
0.531 &
novel
\\
\\

&{\fontsize{9pt}{12pt}\selectfont \textbf{3. } It was an excitement that <11> and I had long dreamed of that scatter of tiny, \textbf{magically} named islands strewn across one third of a globe.} &
0.278 &
conventional 
\\ \\

&{\fontsize{9pt}{12pt}\selectfont \textbf{4. } The seemingly random and <11> designed to disguise a boat's shape from the \textbf{prying} eyes of U-Boat captains, so it <10> in the Bahamas.} &
0.588 &
novel 
 \\ \cmidrule(lr){1-4}

\multirow{4}{*}{News}&
{\fontsize{9pt}{12pt}\selectfont \textbf{5. } One Mr Clarke can not duck away from if he wants to \textbf{avoid} a second Winter of Discontent} &
 -0.094 &
 conventional 
\\ \\

&{\fontsize{9pt}{12pt}\selectfont \textbf{6. } This was conveniently \textbf{encapsulated} in the first try.} &
0.500 &
novel 
\\ \\ 

&{\fontsize{9pt}{12pt}\selectfont \textbf{7. } Thrusts of resistance ( mass demonstrations, resignations, tax rebellions, etc ) would come in \textbf{crests}.} &
0.382 &
conventional 
\\ \\

&{\fontsize{9pt}{12pt}\selectfont \textbf{8. } Travel: A \textbf{pilgrimage} sans progress Elisabeth de Stroumillo potters round Poitou} &
0.514 &
novel 
\\ \cmidrule(lr){1-4}

\multirow{4}{*}{Academic}&
{\fontsize{9pt}{12pt}\selectfont \textbf{9. } The Tehuana dress is by no means the most decorative variant or the \textbf{closest} to pre-Hispanic forms of clothing.} &
 -0.194 &
 conventional 
\\ \\

&{\fontsize{9pt}{12pt}\selectfont \textbf{10. } Interwoven with these images are subtler references to the \textbf{metaphorical} borderlines which separate Latin American <5> and North America.} &
0.529 &
novel 
\\ \\ 

&{\fontsize{9pt}{12pt}\selectfont \textbf{11. } This is often linked with a supposed denunciatory effect — the idea that the mandatory life sentence \textbf{denounces} murder as emphatically as possible <18> this crime.} &
0.294 &
conventional 
\\ \\

&{\fontsize{9pt}{12pt}\selectfont \textbf{12. } He certainly held deep convictions as to the <9>, but at least a part of his apparent \textbf{hostility} was assumed for the occasion, a hard <7> in the end.} &
0.514 &
novel 
\\ \cmidrule(lr){1-4}

\multirow{3}{*}{Conversation}&
{\fontsize{9pt}{12pt}\selectfont \textbf{13. } Me dad said he's had enough Well, we were debating whether to \textbf{give} it to you or not.} &
 -0.633 &
 conventional 
\\ \\

&{\fontsize{9pt}{12pt}\selectfont \textbf{14.} \textbf{Struggled} with it a little} &
0.552 &
novel 
\\ \\ 

&{\fontsize{9pt}{12pt}\selectfont \textbf{15. }That's an old \textbf{trick}.} &
0.310 &
conventional 
\\ \\

&{\fontsize{9pt}{12pt}\selectfont \textbf{16. }Can you \textbf{sort} erm, madame out?} &
0.567 &
novel 
\\

\bottomrule
\end{tabular}
\caption{Examples from the VU Amsterdam Metaphor Corpus (VUAMC). The metaphor word is in \textbf{boldface} within sentences. For simpler presentations, we remove some words from long sentences and replace them with a tag of the number of words removed, e.g. <11>. \textbf{Novelty Score} is the \citet{do-dinh-etal-2018-weeding} normalised human rating, and \textbf{Label} is the binary novelty label based on the 0.5 threshold. Contrastive examples are picked randomly from the dataset for each genre to illustrate the differences between conventional and novel instances according to the ``Novelty Score''.}
\label{tab: data_examples}
\end{table*}

\section{Numerical Results}
\label{sec:appendix_res}
Detailed numerical results of our study are listed in Tables~\ref{tab: results_model_size_metnov}, \ref{tab: results_model_size_surp}, \ref{tab: results_model_step_metnov} and \ref{tab: results_model_step_surp}.

\begin{table}[!htp]
\centering
\small
\begin{tabular}{l | c c c}
\toprule
\textbf{Model} & \textbf{Pearson ($r$)} & \textbf{Spearman ($\rho$)} & \textbf{AUC} \\
\midrule
NLF-Human	 & 		 \textbf{.656} 		 & 		 \textbf{.664} 		 & 		 \textbf{.904} 		 \\
\arrayrulecolor{black!30}\midrule
\arrayrulecolor{black}
NLF-LM 		 & 		 \textbf{.599} 		 & 		 \textbf{.633} 		 & 		 \textbf{.899} 		 \\
\arrayrulecolor{black!30}\midrule
\arrayrulecolor{black}
Pythia-70M 		 & 		 \textbf{.448} 		 & 		 \textbf{.453} 		 & 		 .832 		 \\
Pythia-160M 		 & 		 .426 		 & 		 .425 		 & 		 \textbf{.833} 		 \\
Pythia-410M 		 & 		 .382 		 & 		 .374 		 & 		 .792 		 \\
Pythia-1B 		 & 		 .371 		 & 		 .359 		 & 		 .782 		 \\
Pythia-1.4B 		 & 		 .357 		 & 		 .344 		 & 		 .771 		 \\
Pythia-2.8B 		 & 		 .351 		 & 		 .336 		 & 		 .766 		 \\
Pythia-6.9B 		 & 		 .338 		 & 		 .322 		 & 		 .758 		 \\
Pythia-12B 		 & 		 .330 		 & 		 .312 		 & 		 .752 		 \\
\bottomrule
\end{tabular}
\caption{Pearson's $r$ and Spearman's $\rho$ Correlation and AUC estimates between \textbf{Metaphor Novelty Scores} and \textbf{Surprisal}; \textbf{Negative Log Word Frequency in general language use (NLF-Human)}; and \textbf{Negative Log Word Frequency in Pythia's pretraining data (NLF-LM)} across different model sizes. All reported estimates are significant at the 0.001 level.}
\label{tab: results_model_size_metnov}
\end{table}

\begin{table}[!htp]
\centering
\small
\begin{tabular}{l | c c}
\toprule
\textbf{Model} & \textbf{NLF-Human ($\rho$)} & \textbf{NLF-LM ($\rho$)} \\
\midrule
Pythia-70M 		 & 		 \textbf{.572} 		 & 		 \textbf{.606} \\
Pythia-160M 		 & 		 .533 		 & 		 .573 \\
Pythia-410M 		 & 		 .467 		 & 		 .507 \\
Pythia-1B 		 & 		 .449 		 & 		 .489 \\
Pythia-1.4B 		 & 		 .434 		 & 		 .474 \\
Pythia-2.8B 		 & 		 .421 		 & 		 .462 \\
Pythia-6.9B 		 & 		 .404 		 & 		 .445 \\
Pythia-12B 		 & 		 .396 		 & 		 .438 \\
\bottomrule
\end{tabular}
\caption{Spearman Correlation estimates between \textbf{Surprisal} and \textbf{Negative Log Word Frequency in general language use (NLF-Human)}; and \textbf{Negative Log Word Frequency in Pythia's pretraining data (NLF-LM)} across different model sizes. All reported estimates are significant at the 0.001 level.}
\label{tab: results_model_size_surp}
\end{table}

\begin{table*}[!htp]
\centering
\small
\begin{tabular}{l c | c c | c c}
\toprule
 \textbf{\# Steps} & \textbf{\# Pretraining Tokens} & \multicolumn{2}{c}{\textbf{Surprisal}}  & \multicolumn{2}{|c}{\textbf{NLF-LM}} \\
\midrule
 &  & \textbf{$\rho$} & AUC & $\rho$ &  AUC \\
\midrule
1 	 & 	 2M 	 & 	 .221 	 & 	 .799 	 & 	 .626 	 & 	 .891 \\
2 	 & 	 4M 	 & 	 .221 	 & 	 .799 	 & 	 .631 	 & 	 .894 \\
4 	 & 	 8M 	 & 	 .221 	 & 	 .799 	 & 	 .631 	 & 	 .896 \\
8 	 & 	 17M 	 & 	 .224 	 & 	 .800	 & 	 .631 	 & 	 .897 \\
16 	 & 	 34M 	 & 	 .226 	 & 	 .799 	 & 	 .631 	 & 	 .898 \\
32 	 & 	 67M 	 & 	 .250 	 & 	 .804 	 & 	 .632 	 & 	 .898 \\
64 	 & 	 134M 	 & 	 .435 	 & 	 .858 	 & 	 .632 	 & 	 \textbf{.899} \\
128 	 & 	 268M 	 & 	 \textbf{.618} 	 & 	 \textbf{.895} 	 & 	 .632 	 & 	 .899 \\
256 	 & 	 537M 	 & 	 .574 	 & 	 .893 	 & 	 \textbf{.633} 	 & 	 .899 \\
512 	 & 	 1B 	 & 	 .554 	 & 	 .888 	 & 	 .633 	 & 	 .899 \\
1,000 	 & 	 2B 	 & 	 .524 	 & 	 .873 	 & 	 .633 	 & 	 .899 \\
2,000 	 & 	 4B 	 & 	 .473 	 & 	 .849 	 & 	 .633 	 & 	 .899 \\
3,000 	 & 	 6B 	 & 	 .475 	 & 	 .846 	 & 	 .633 	 & 	 .899 \\
4,000 	 & 	 8B 	 & 	 .458 	 & 	 .842 	 & 	 .633 	 & 	 .899 \\
5,000 	 & 	 10B 	 & 	 .458 	 & 	 .842 	 & 	 .633 	 & 	 .899 \\
6,000 	 & 	 13B 	 & 	 .459 	 & 	 .841 	 & 	 .633 	 & 	 .899 \\
7,000 	 & 	 15B 	 & 	 .457 	 & 	 .836 	 & 	 .633 	 & 	 .899 \\
8,000 	 & 	 17B 	 & 	 .453 	 & 	 .833 	 & 	 .633 	 & 	 .899 \\
9,000 	 & 	 19B 	 & 	 .451 	 & 	 .833 	 & 	 .633 	 & 	 .899 \\
12,000 	 & 	 25B 	 & 	 .459 	 & 	 .838 	 & 	 .633 	 & 	 .899 \\
17,000 	 & 	 36B 	 & 	 .448 	 & 	 .832 	 & 	 .633 	 & 	 .899 \\
22,000 	 & 	 46B 	 & 	 .447 	 & 	 .831 	 & 	 .633 	 & 	 .899 \\
27,000 	 & 	 57B 	 & 	 .446 	 & 	 .836 	 & 	 .633 	 & 	 .899 \\
32,000 	 & 	 67B 	 & 	 .453 	 & 	 .835 	 & 	 .633 	 & 	 .899 \\
37,000 	 & 	 78B 	 & 	 .446 	 & 	 .839 	 & 	 .633 	 & 	 .899 \\
42,000 	 & 	 88B 	 & 	 .450 	 & 	 .843 	 & 	 .633 	 & 	 .899 \\
47,000 	 & 	 99B 	 & 	 .445 	 & 	 .833 	 & 	 .633 	 & 	 .899 \\
52,000 	 & 	 109B 	 & 	 .444 	 & 	 .829 	 & 	 .633 	 & 	 .899 \\
57,000 	 & 	 120B 	 & 	 .443 	 & 	 .827 	 & 	 .633 	 & 	 .899 \\
62,000 	 & 	 130B 	 & 	 .442 	 & 	 .831 	 & 	 .633 	 & 	 .899 \\
67,000 	 & 	 141B 	 & 	 .446 	 & 	 .831 	 & 	 .633 	 & 	 .899 \\
72,000 	 & 	 151B 	 & 	 .441 	 & 	 .830 	 & 	 .633 	 & 	 .899 \\
77,000 	 & 	 161B 	 & 	 .447 	 & 	 .828 	 & 	 .633 	 & 	 .899 \\
82,000 	 & 	 172B 	 & 	 .445 	 & 	 .824 	 & 	 .633 	 & 	 .899 \\
87,000 	 & 	 182B 	 & 	 .440 	 & 	 .826 	 & 	 .633 	 & 	 .899 \\
92,000 	 & 	 193B 	 & 	 .442 	 & 	 .831 	 & 	 .633 	 & 	 .899 \\
97,000 	 & 	 203B 	 & 	 .443 	 & 	 .826 	 & 	 .633 	 & 	 .899 \\
102,000 	 & 	 214B 	 & 	 .449 	 & 	 .828 	 & 	 .633 	 & 	 .899 \\
107,000 	 & 	 224B 	 & 	 .448 	 & 	 .829 	 & 	 .633 	 & 	 .899 \\
112,000 	 & 	 235B 	 & 	 .453 	 & 	 .833 	 & 	 .633 	 & 	 .899 \\
117,000 	 & 	 245B 	 & 	 .447 	 & 	 .828 	 & 	 .633 	 & 	 .899 \\
122,000 	 & 	 256B 	 & 	 .455 	 & 	 .835 	 & 	 .633 	 & 	 .899 \\
127,000 	 & 	 266B 	 & 	 .451 	 & 	 .830 	 & 	 .633 	 & 	 .899 \\
132,000 	 & 	 277B 	 & 	 .451 	 & 	 .835 	 & 	 .633 	 & 	 .899 \\
137,000 	 & 	 287B 	 & 	 .448 	 & 	 .829 	 & 	 .633 	 & 	 .899 \\
143,000 	 & 	 300B 	 & 	 .453 	 & 	 .832 	 & 	 .633 	 & 	 .899 \\
\bottomrule
\end{tabular}
\caption{Spearman Correlation and AUC estimates between \textbf{Metaphor Novelty Scores} and \textbf{Surprisal}; and \textbf{Negative Log Word Frequency in Pythia's pretraining data (NLF-LM)} across pretraining steps of Pythia-70M, reporting the amount of pretraining data tokens seen at each step.}
\label{tab: results_model_step_metnov}
\end{table*}

\begin{table*}[!htp]
\centering
\small
\begin{tabular}{l c | c c}
\toprule
\textbf{\# Steps} & \textbf{\# Pretraining Tokens} & \textbf{NLF-Human ($\rho$)} & \textbf{NLF-LM ($\rho$)} \\
\midrule
1 	 & 	 2M 	 & 	 .286 	 & 	 .302 \\
2 	 & 	 4M 	 & 	 .286 	 & 	 .306 \\
4 	 & 	 8M 	 & 	 .286 	 & 	 .309 \\
8 	 & 	 17M 	 & 	 .292 	 & 	 .316 \\
16 	 & 	 34M 	 & 	 .305 	 & 	 .332 \\
32 	 & 	 67M 	 & 	 .352 	 & 	 .388 \\
64 	 & 	 134M 	 & 	 .623 	 & 	 .679 \\
128 	 & 	 268M 	 & 	 \textbf{.885} 	 & 	 \textbf{.953} \\
256 	 & 	 537M 	 & 	 .792 	 & 	 .866 \\
512 	 & 	 1B 	 & 	 .730 	 & 	 .778 \\
1,000 	 & 	 2B 	 & 	 .679 	 & 	 .705 \\
2,000 	 & 	 4B 	 & 	 .597 	 & 	 .629 \\
3,000 	 & 	 6B 	 & 	 .601 	 & 	 .634 \\
4,000 	 & 	 8B 	 & 	 .580 	 & 	 .615 \\
5,000 	 & 	 10B 	 & 	 .587 	 & 	 .623 \\
6,000 	 & 	 13B 	 & 	 .583 	 & 	 .609 \\
7,000 	 & 	 15B 	 & 	 .579 	 & 	 .614 \\
8,000 	 & 	 17B 	 & 	 .578 	 & 	 .606 \\
9,000 	 & 	 19B 	 & 	 .572 	 & 	 .609 \\
12,000 	 & 	 25B 	 & 	 .583 	 & 	 .614 \\
17,000 	 & 	 36B 	 & 	 .572 	 & 	 .605 \\
22,000 	 & 	 46B 	 & 	 .566 	 & 	 .595 \\
27,000 	 & 	 57B 	 & 	 .566 	 & 	 .599 \\
32,000 	 & 	 67B 	 & 	 .578 	 & 	 .612 \\
37,000 	 & 	 78B 	 & 	 .563 	 & 	 .602 \\
42,000 	 & 	 88B 	 & 	 .570 	 & 	 .604 \\
47,000 	 & 	 99B 	 & 	 .564 	 & 	 .600 \\
52,000 	 & 	 109B 	 & 	 .562 	 & 	 .598 \\
57,000 	 & 	 120B 	 & 	 .559 	 & 	 .595 \\
62,000 	 & 	 130B 	 & 	 .557 	 & 	 .591 \\
67,000 	 & 	 141B 	 & 	 .560 	 & 	 .595 \\
72,000 	 & 	 151B 	 & 	 .559 	 & 	 .597 \\
77,000 	 & 	 161B 	 & 	 .565 	 & 	 .601 \\
82,000 	 & 	 172B 	 & 	 .563 	 & 	 .597 \\
87,000 	 & 	 182B 	 & 	 .557 	 & 	 .595 \\
92,000 	 & 	 193B 	 & 	 .557 	 & 	 .592 \\
97,000 	 & 	 203B 	 & 	 .557 	 & 	 .594 \\
102,000 	 & 	 214B 	 & 	 .565 	 & 	 .597 \\
107,000 	 & 	 224B 	 & 	 .563 	 & 	 .597 \\
112,000 	 & 	 235B 	 & 	 .571 	 & 	 .604 \\
117,000 	 & 	 245B 	 & 	 .562 	 & 	 .595 \\
122,000 	 & 	 256B 	 & 	 .572 	 & 	 .607 \\
127,000 	 & 	 266B 	 & 	 .568 	 & 	 .602 \\
132,000 	 & 	 277B 	 & 	 .567 	 & 	 .602 \\
137,000 	 & 	 287B 	 & 	 .563 	 & 	 .600 \\
143,000 	 & 	 300B 	 & 	 .572 	 & 	 .606 \\
\bottomrule
\end{tabular}
\caption{Spearman Correlation estimates between \textbf{Surprisal} and \textbf{Negative Log Word Frequency in general corpora}; and \textbf{Negative Log Word Frequency in pretraining corpora} across pretraining steps of Pythia-70M, reporting the amount of pretraining data tokens seen at each step.}
\label{tab: results_model_step_surp}
\end{table*}

\end{document}